\documentclass[sigconf]{acmart}
\usepackage{amsmath,amsfonts,amsthm}
\usepackage{algorithmic}
\usepackage{graphicx}
\usepackage{textcomp}
\usepackage{booktabs}
\usepackage{url}
\usepackage{multirow}
\usepackage{balance}

\AtBeginDocument{%
 \providecommand\BibTeX{{%
  \normalfont B\kern-0.5em{\scshape i\kern-0.25em b}\kern-0.8em\TeX}}}

\copyrightyear{2021} 
\acmYear{2021} 
\setcopyright{acmcopyright}\acmConference[CIKM '21]{Proceedings of the 30th ACM International Conference on Information and Knowledge Management}{November 1--5, 2021}{Virtual Event, QLD, Australia}
\acmBooktitle{Proceedings of the 30th ACM International Conference on Information and Knowledge Management (CIKM '21), November 1--5, 2021, Virtual Event, QLD, Australia}
\acmPrice{15.00}
\acmDOI{10.1145/3459637.3482464}
\acmISBN{978-1-4503-8446-9/21/11}

\begin{document}
\fancyhead{}
\title{Power to the Relational Inductive Bias: Graph Neural Networks in Electrical Power Grids}

\author{Martin Ringsquandl}

\authornote{Both authors contributed equally to this research.}
\affiliation{%
 \institution{Siemens \country{Germany}}}
 \email{martin.ringsquandl@siemens.com}

\author{Houssem Sellami}
\authornotemark[1]
\affiliation{%
 \institution{Siemens\country{Germany}}}
 \email{houssem.sellami@siemens.com}
 
\author{Marcel Hildebrandt}
\affiliation{%
 \institution{Siemens  \country{Germany}}}
 \email{marcel.hildebrandt@siemens.com}
 
\author{Dagmar Beyer}
\affiliation{%
 \institution{Siemens  \country{Germany}}
}
 \email{dagmar.beyer@siemens.com}
 
\author{Sylwia Henselmeyer}
\author{Sebastian Weber}
\affiliation{%
 \institution{Siemens   \country{Germany}}}
 \email{sylwia.henselmeyer@siemens.com}
 
\author{Mitchell Joblin}
\affiliation{%
 \institution{Siemens \country{Germany}}}
 \email{mitchell.joblin@siemens.com}
\renewcommand{\shortauthors}{Ringsquandl and Sellami, et al.}

\begin{abstract}
The application of graph neural networks (GNNs) to the domain of electrical power grids has high potential impact on smart grid monitoring. Even though there is a natural correspondence of power flow to message-passing in GNNs, their performance on power grids is not well-understood. We argue that there is a gap between GNN research driven by benchmarks which contain graphs that differ from power grids in several important aspects. Additionally, inductive learning of GNNs across multiple power grid topologies has not been explored with real-world data. 
We address this gap by means of (i) defining power grid graph datasets in inductive settings, (ii) an exploratory analysis of graph properties, and (iii) an empirical study of the concrete learning task of state estimation on real-world power grids.
Our results show that GNNs are more robust to noise with up to 400\% lower error compared to baselines.
Furthermore, due to the unique properties of electrical grids, we do not observe the well known over-smoothing phenomenon of GNNs and find the best performing models to be exceptionally deep with up to 13 layers. This is in stark contrast to existing benchmark datasets where the consensus is that 2--3 layer GNNs perform best. Our results demonstrate that a key challenge in this domain is to effectively handle long-range dependence.


\end{abstract}




\maketitle


\section{Introduction}
Graph neural networks (GNNs) are an increasingly popular class of machine learning (ML) models with many potential industrial applications.
GNNs have been successfully applied to various problems that deal with graph-structured data, such as classification of social networks or chemical compounds \cite{Fan2019,Zitnik2018}, forecasting flows in traffic networks \cite{Chen2019}, as well as relational reasoning \cite{Schlichtkrull2018}.
In general, ML on power grids is considered a major contributor to the emergence of smart grids facilitating the monitoring and control of critical infrastructure with increasing power volatility introduced by more decentralized energy production \cite{Cheng2019}.
In GNN research, electrical power grids are still a largely unexplored application domain, although their graph representations and flow algorithms have been studied extensively \cite{Pagani2013}.
Intuitively, the principles of physical power flow in such grids is a natural fit to the message-passing strategy used by GNNs, similar to inference in graphical models which have been used for power grids \cite{Weng2013}. A more popular way to describe the GNN message-passing in engineering domains is the learning of locally-bound spectral filters, hence exploiting spatial smoothness of graph signals (e.g. measurements on nodes). A powerful feature of GNNs is their inductive and transfer learning capabilities, analogous to convolutional neural networks (CNNs) on image data. This opens up new key use cases for power grid operators, where models can be trained on a set of different grid topologies and applied to novel ones. One such use case is power grid monitoring using trained GNNs as surrogate for physics-based power flow solvers. GNNs are as other learning-based approaches less sensitive to noise and inference is much faster compared to traditional solvers \cite{Donon2019}. Despite the GNNs' powerful inductive capability, most state-of-the-art ML models in the power grid domain (e.g. for state estimation) are typically dedicated to a single grid topology with no transfer possibilities. No transfer also means that these models need to be re-trained in case their dedicated topology changes.
In this paper, we explore both the opportunities as well as challenges of applying GNNs in the power grid domain.
We focus on an inductive setting where models are both trained and deployed on various different power grid topologies.


\begin{figure*}[h]
 \centering
 \includegraphics[width=\textwidth]{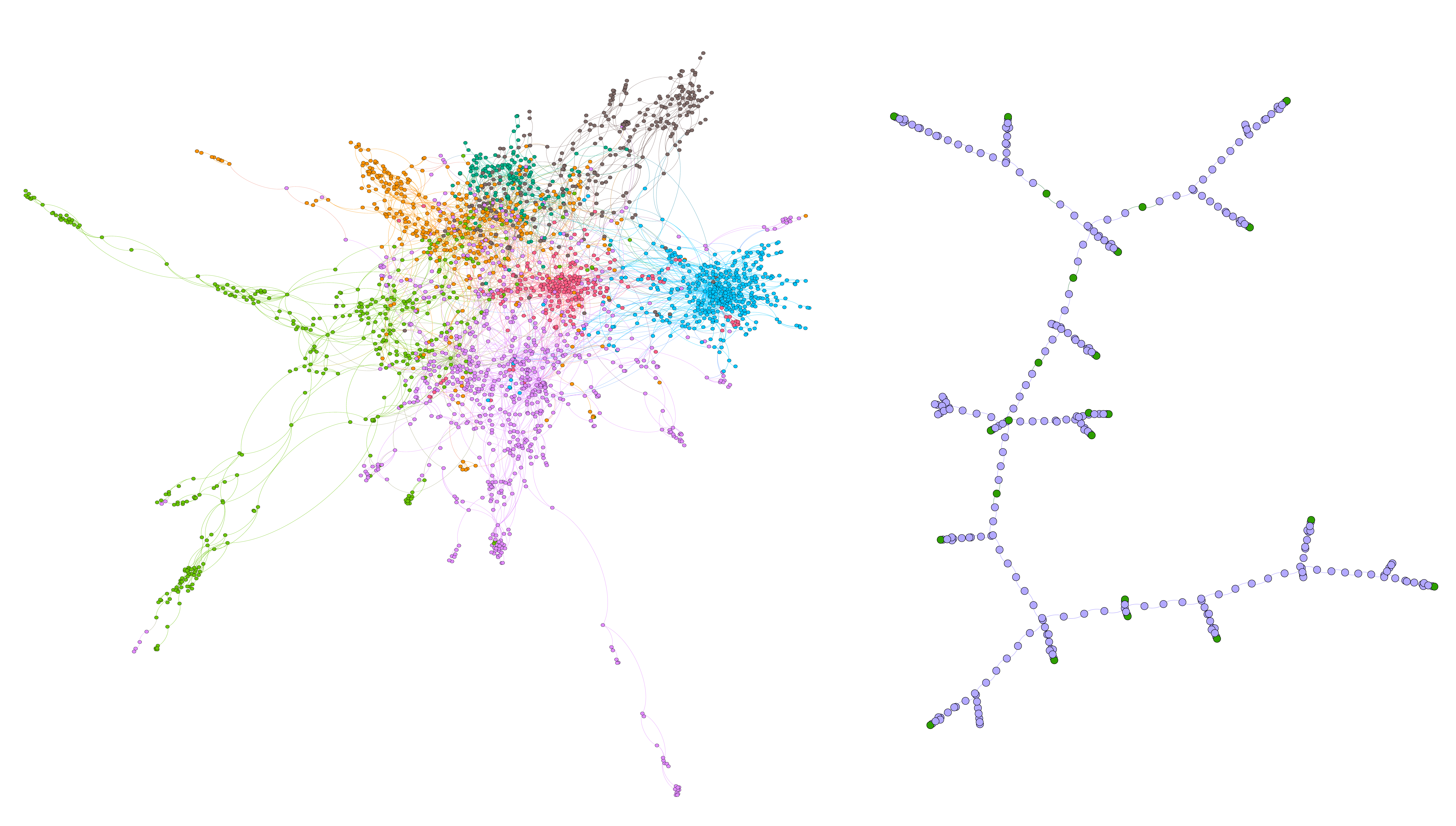}
 \caption{Topology comparison. Left: Cora benchmark citation network, Right: Distribution power grid. Same layouting algorithm was used for both plots.}
 \label{fig:teaser_figure}
\end{figure*}

In Figure~\ref{fig:teaser_figure} a comparison between the Cora benchmark citation network on the left and an electrical distribution power grid on the right is shown. Already a simple visual comparison gives an impression about the differences in structural properties. The colors for Cora denote  nodes' class labels. In the power grid green nodes contain measurements, such as power, voltage and current, which can be seen as noisy labels, the rest is unlabeled. It is apparent that a rather shallow GNN model (with a limited receptive field) suffices to exploit the local community structure of the Cora network, while it would not be able to effectively propagate information due to label sparseness and non-existing clustering in the power grid.	

The Open Graph Benchmark (OGB)~\cite{hu2020open} is a recent effort to bring more diverse and larger datasets into the research community. Electrical power grids, despite their industrial relevance, are not included as of today. Interestingly, power grids possess unique structural characteristics (e.g., low clustering, high diameter, and long-range dependence) as well as node features (e.g., noisy and sparse measurements and heterogeneous node types). The omission of electrical grids from the OGB represents a significant barrier to understanding the limitations of existing GNNs and developing new architectures and training techniques to overcome the challenges presented by this real-world industrial problem setting. 




Addressing this gap in GNN research, we make the following contributions:
\begin{itemize}
  \item We discuss open challenges concerning the application of GNNs to power grids and limitations of existing ML approaches.
  \item We show that power grids contain unique topological features and partially disjoint node attributes making them significantly different from standard benchmark datasets.
  \item We conduct an empirical study showing that under multiple scenarios with real-world power grids that well-known GNN models fail to perform adequately on the state estimation task.
  \item We propose a power grid data model suitable for inductive learning and exceptionally deep GNN models to handle long-range dependence introduced by physical power flows by efficiently aggregating node representations from very distant neighborhoods. 
  \item We show that certain GNN architectures are more robust to noisy labels and are capable of enabling a new key use case, generalizing to novel topologies, ultimately allowing power grid operators to deploy more accurate and robust ML models.
\end{itemize}


\section{Problem Setup}
In this section, we give a brief description of power grid data and formalize the state estimation problem as an ML task. This task is used as a canonical example throughout the paper.

\paragraph{Power Grid Data}
For our purpose a power grid is represented by an undirected graph $\mathcal{G} =(\mathcal{V}, \mathcal{E})$, where $\mathcal{V} = \{ v_1, v_2, \dots, v_n \}$ denotes the indexed vertex set and $\mathcal{E}$ the edge set. Vertices correspond to physical components that comprise the power grid. Each component has an equipment type in \sloppy $\mathcal{T} = \{\text{Generator}, \text{Transformer}, \text{Busbar}, \dots \}$ inducing a type assignment function $f_\text{type}: \mathcal{V} \rightarrow \mathcal{T}$ such that $f_\text{type}(v)$ indicates the equipment type of vertex $v$. To ease the notion we denote with $\mathcal{V}_T := f^{-1}(T) $ the set of vertices of type $T$. For example, $\mathcal{V}_\text{Busbar}$ denotes the set of all busbars in $\mathcal{G}$. Whenever there exists a direct physical connection between two components, there is an edge in $\mathcal{E}$ linking the corresponding vertices. 

We assign a set of $d \in \mathbb{N}$ static features to each vertex that represent time-invariant characteristics of the underlying component. Examples of static features include the nominal voltage, wiring types, or reactance of a component.
The entirety of all these features is captured in the design matrix $\mathbf{C} \in \mathbb{R}^{n \times d}$. Throughout this work, we use elements of indexed sets and their indices interchangeably (e.g., the $C_{v_i}$ and $C_i$ equivalently denote the static features of $v_i$). In case a feature is not applicable to certain component types (e.g., the type of wiring is a characteristic of a transformers only), we apply zero padding. 

The dynamic state\footnote{For simplicity we assume that in case of three power phases that they are in balance and we can ignore phase angles} of a power grid $\mathcal{G}$ at time $t$ is given by the current $Z_{v,I}^{(t)} \in \mathbb{R}$, the active power $Z_{v,P}^{(t)} \in \mathbb{R}$, the re-active power $Z_{v,Q}^{(t)} \in \mathbb{R}$, and the voltage $Z_{v,V}^{(t)} \in \mathbb{R}$ at each vertex $v \in \mathcal{V}$. 
More compactly, the state of $\mathcal{G}$ at time $t$ is determined by the matrix $Z^{(t)} \in \mathbb{R}^{n \times 4}$, where the $i$-th row is given by $Z^{(t)}_{i} = \left[Z_{v_i,I}^{(t)}, Z_{v_i,P}^{(t)}, Z_{v_i,Q}^{(t)}, Z_{v_i,V}^{(t)}\right]$. 

In most real-world applications, the dynamic state of $\mathcal{G}$ is monitored by equipping components with measurement sensors. Due to high costs of installing such sensors, the resulting observations are typically very sparse. For example, in distribution power grids, the percentage of equipment that has at least one observation at a given time is often less than 10\%. 

\paragraph{State Estimation Problem}
One of the most studied ML tasks on power grids is state estimation (SE). SE aims to approximate the voltages and phase angles at junction nodes, corresponding to busbars, of a power system by propagating information of injections and measurements across the system’s network. More formally, suppose that for any point in time $t$, the index set $\mathcal{I}_t \subset \{1, 2, \dots, n\} \times \{I,P,Q,V\}$ indicates the set of measurements taken at time $t$. This means, neglecting measurement errors, for $(i,j) \in \mathcal{I}_t$ the quantity $Z^{(t)}_{i,j}$ is observed. For convenience, we omit the time index and denote with $Z_{\mathcal{I}_t}$ the partially observed state of $\mathcal{G}$ at $t$. This leads to the following definition.

\begin{definition}[State estimation (SE)]
Given a power grid $\mathcal{G}=(\mathcal{V}, \mathcal{E})$ and a set of measurements $\mathcal{I}_t$, the task of state estimation consists of approximating the voltages of all busbars denoted by $Z^{(t)}_{\mathcal{V}_\text{Busbar}, V}$. That means we aim to infer $Z^{(t)}_{\mathcal{V}_\text{Busbar}, V}$ from the partially observed states $Z_{\mathcal{I}_t}$.
\end{definition}
Note that the partially observed states and the busbar voltages are typically disjoint. As we will show in section \ref{sec:gnn}, given this graph data representation, SE corresponds to a semi-supervised node regression task. 
On the other hand, traditionally, SE was addressed by building a system of equations 
\begin{equation}
\label{eq:se}
  Z_{\mathcal{I}_t} = f(Z^{(t)}_{\mathcal{V}_\text{Busbar}, V}) + \mathbf{\epsilon}_t \, ,
\end{equation}
where $f(\cdot)$ is a non-linear function relating the hidden (voltage) state vectors to the measurements. $\epsilon_t$ is a vector of measurement noise. Note that these equations implicitly encode the power grid topology, since the system of equations are taking the busbar connections into account.
Also, due to the non-linear power equations, there exists in general no closed-form solution to \eqref{eq:se}. A popular method used in practice is a weighted-least-squares (WLS) fit of $Z^{(t)}_{\mathcal{V}_\text{Busbar}, V}$ with a linearity assumption imposed on $f(\cdot)$, which then requires iterative fitting until convergence. Therefore, WLS produces a global fit of the states to minimize the error w.r.t. the measurements. 

\paragraph{Euclidean Neural Network Formulation}
When using neural network (NN) models for SE, the weights cannot directly represent the state variables to be fitted. Instead the sides of Eq.~\ref{eq:se} are swapped and the NN weights $\mathbf{W}$ transform partial observations into the busbar states of interest. 
The partial observations $Z_{\mathcal{I}_t}$ at each point in time $t$ are flattened into a vector (neglecting topology information) $\mathbf{z}_t \in \mathcal{R}^{4n}$ and the state estimation is phrased as a multi-output regression model

\begin{equation}
\label{eq:nn}
  \hat{Z}^{(t)}_{\mathcal{V}_\text{Busbar}, V} = f_{NN}(\mathbf{z}_t; \mathbf{W}) \, .
\end{equation}

A natural extension to this formulation is instead of independently fitting a NN to each time-dependent state to use recurrent neural networks (RNN) which can exploit state and measurement smoothness across time in a sliding window of $(m+1)$ time steps.
\begin{equation}
\label{eq:segnn}
  \hat{Z}^{(t+1)}_{\mathcal{V}_\text{Busbar}, V} = f_{RNN}(\mathbf{z}_{t+1}, \mathbf{z}_{t}, \dots, \mathbf{z}_{t-m} ; \mathbf{W}) \, .
\end{equation}

Having ground-truth states as regression labels of all busbars, usually obtained from simulations, we can define the mean-squared error objective as following:
\begin{equation}
\label{eq:mse}
	\mathcal{L} = \frac{1}{|\mathcal{V}_\text{Busbar}|}\sum_{v \in \mathcal{V}_\text{Busbar}}\left( Z^{(t)}_{v, V} -\hat{Z}^{(t)}_{v, V}\right)^2 \, .
\end{equation}



A major weakness of all the above mentioned models is that they are bound to a single topology because the model is only aware of the measurement vector as input. Hence, the NN implicitly trains its weights to account for the spatial connections in the input. In this paper, we propose to explicitly incorporate the topology by means of graph-based ML models and explore the benefits this offers. 

\begin{figure}
 \centering
 \includegraphics[width=\linewidth]{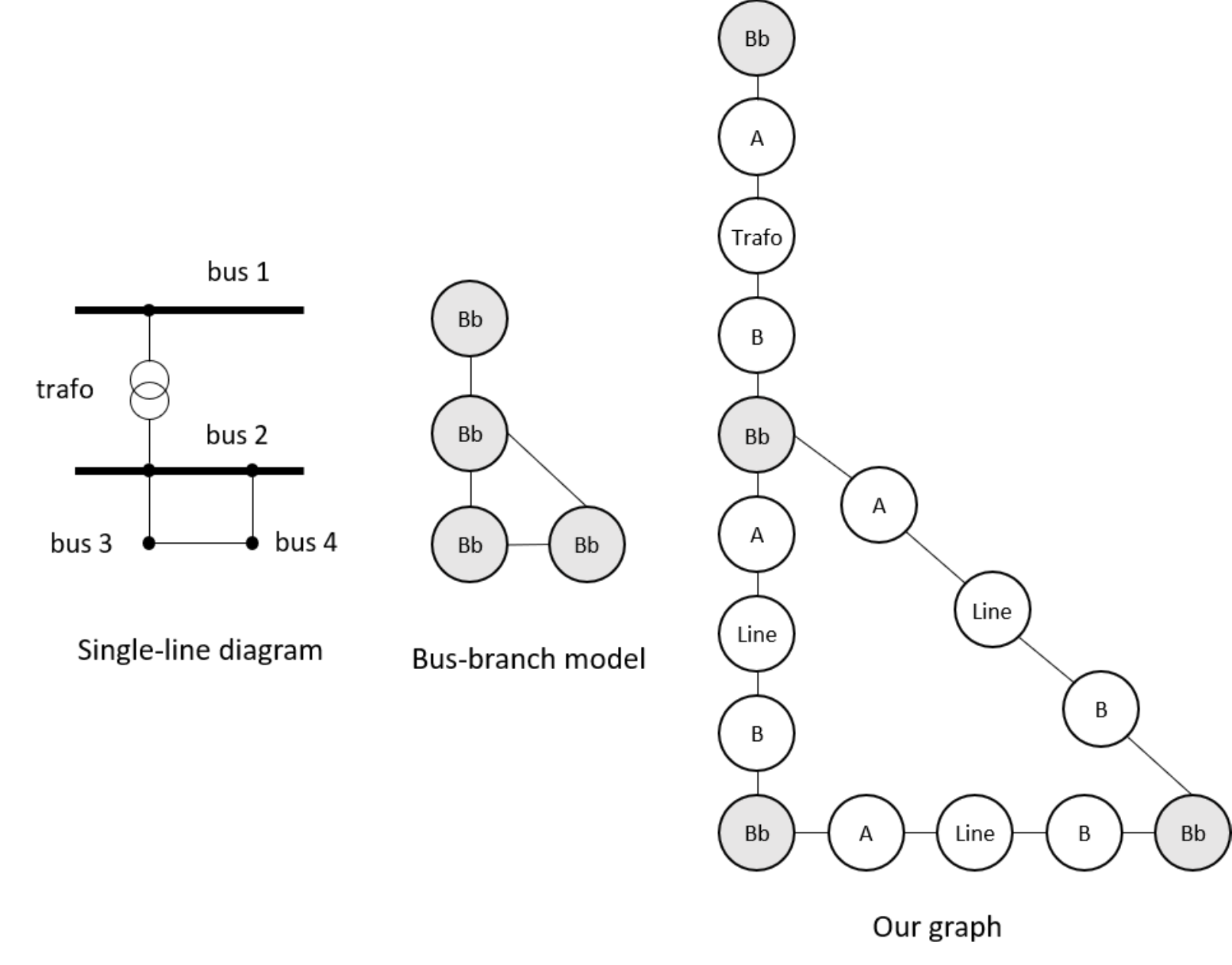}
 \caption{Comparison of power grid graph representations. Left: Single-line diagram notation, middle:bus-branch model, right: our proposal for inductive learning}
 \label{fig:bus_branch}
\end{figure}

\paragraph{Graph representations for ML}
There are several approaches on how to represent power grid topologies as graphs. First, electrical engineers typically use the \textit{single-line-diagram} as a graphical notation, involving different symbols for types of equipment. 
A frequently used data model abstraction is the so-called \textit{bus-branch} model for topology representation in which only busbars are considered as nodes and any connecting equipment between busbars is modeled as an edge (branch). 
Going back to our definition of $\mathcal{G}$ this would mean only busbars are part of $\mathcal{V}$, any other equipment is represented within the edge set $\mathcal{E}$.
In Figure~\ref{fig:bus_branch} we show an example topology in three different representations: single-line diagram, bus-branch model, and our graph model. 

The bus-branch representations has two fundamental limitations: 
\begin{itemize}
  \item \textbf{Aliasing of branch equipment connections}: Any branch equipment needs to be modeled as edge attributes (features) thereby losing information on branch equipment connections.
  \item \textbf{Missing equipment extremities}: Measurements taken on lines or transformers need to specify the exact side (origin or extremity) where they were taken. This cannot be represented in the bus-branch, unless resorting to inventing edge attributes that express this implicitly.
\end{itemize}
A simplifying assumption that is often made in ML approaches with the bus-branch model is that equipment on the branches have identical static attributes. For example, all power lines are assumed to have the same length and conductance.

Because we focus on inductive learning across different topologies, it is crucial to resolve these topology representation shortcomings. Similar to the prior GNN for power flow work \cite{Donon2019}, we developed a more expressive graph representation.
In our case every equipment corresponds to a node. In addition, we introduce two special nodes representing the origin (A) and extremity (B) of a line (low-voltage/high-voltage side of transformers). Formally, $\mathcal{T}' = \mathcal{T} \cup \{ \text{A}, \text{B} \}$, and for every non-busbar equipment $v_i$ on a branch we introduce two additional edges $(v_i, v_{i+1})$, $(v_i, v_{i+2})$ with $f_{\text{type}}(v_{i+1}) = \text{A}$ and $f_{\text{type}}(v_{i+2}) = \text{B}$, while reconnecting the busbars to the A,B nodes, respectively. If multiple equipment lie on a branch, we chain the A,B nodes together and only reconnect the outermost A,B nodes to the busbars.
This allows us to represent all static features $\mathbf{C}$ of equipment as node features as well as measurements on A,B nodes (dynamic state $\mathbf{Z}_t$).

The downside of this more expressive graph representation is that the path length between all nodes increases. Dealing with these long-range dependencies is a challenge that we will address in section~\ref{sec:gnn}. An analysis of the statistical properties of these augmented topologies follows in section~\ref{sec:results}.

\section{Graph neural network model for SE}
\label{sec:gnn}
In this section we translate the standard SE problem to GNNs as semi-supervised node regression. Given a power grid $\mathcal{G}$ we define a generic GNN architecture with $K$ layers and a node regression output as:
\begin{equation}
  \begin{split}
    \mathbf{H}^{(k+1)} &= \sigma(Agg_{\mathcal{E}}(\mathbf{W}^{(k)} \mathbf{H}^{(k)})) \\
     \hat{Z}^{(t)}_{\mathcal{V}_\text{Busbar}, V} &= \mathbf{H}^{(K)}_{[\mathcal{V}_\text{Busbar}]}\mathbf{w}_{out} 
  \end{split}
\end{equation}

where $\mathbf{H}^{(0)} = \mathbf{Z}_t \Vert \mathbf{C}$ is the concatenated node feature matrix of static and dynamic attributes and $Agg_{{\mathcal{E}}}$ represents a commutative aggregation function that pools information of each equipment node's neighborhood defined in $\mathcal{E}$. Commonly used aggregation functions are mean, sum, and maximum pooling. Each $\mathbf{H}^{(k)} \in \mathcal{R}^{n \times d_k}$ contains the $k$-the layer node embeddings. We use the square bracket notation to sub-index $\mathbf{H}^{(k)}$ to the set of busbar nodes.
The vector-valued output in this case is given by a linear transform of the last layers' embeddings with $\mathbf{w}_{out}$ to obtain the state estimates for all busbar nodes. This model can now be trained according to the objective function in Eq.~\ref{eq:mse}. Note that this will produce a local fit which is fundamentally different from the global WLS.

By definition, aggregating information across paths of length $K$ in the topology requires stacking $K$ layers in the GNN architecture. The problems arising with deep GNNs are still actively researched \cite{alon2021bottleneck}. One interpretation is that stacking many layers usually leads to exponentially growing receptive fields which over-smooth information from longer paths. Some works have been proposed trying to overcome this issue, including graph attention (GAT) \cite{velickovic2018} and jumping knowledge \cite{Xu2018}. As we will describe later, the receptive fields in power grid graphs do not have exponential growth, but are still very susceptible to \textit{under-reaching}, meaning the GNNs receptive field need to include long-range dependence.

\paragraph{The Jumping Knowledge architecture}
The jumping knowledge (JK) model is an extension that leverages intermediate node representations from all layers instead of just the one for the final classification or regression task. In JK, the main idea is to connect for each node $v$ all intermediate layer embeddings $\mathbf{h}_v$ to the last layer and use a general layer-aggregation layer to build the target output. In the original version the authors used an aggregation module based on concatenation $\mathbf{h}_v^{(1)} \, \Vert \, \ldots \, \Vert \, \mathbf{h}_v^{(K)}$, max pooling $\max \left( \mathbf{h}_v^{(1)}, \ldots, \mathbf{h}_v^{(K)} \right)$ or weighted summation $\sum_{k=1}^K \alpha_v^{(k)} \mathbf{h}_v^{(k)}$ with attention scores $\alpha_v^{(k)}$ obtained from a forward-LSTM and a backward-LSTM. In the last option, which is a node adaptive mechanism, each node selects the most useful neighborhood ranges based on $\alpha_v$.
\begin{figure*}[t]
 \centering
 \includegraphics[width=0.82\textwidth]{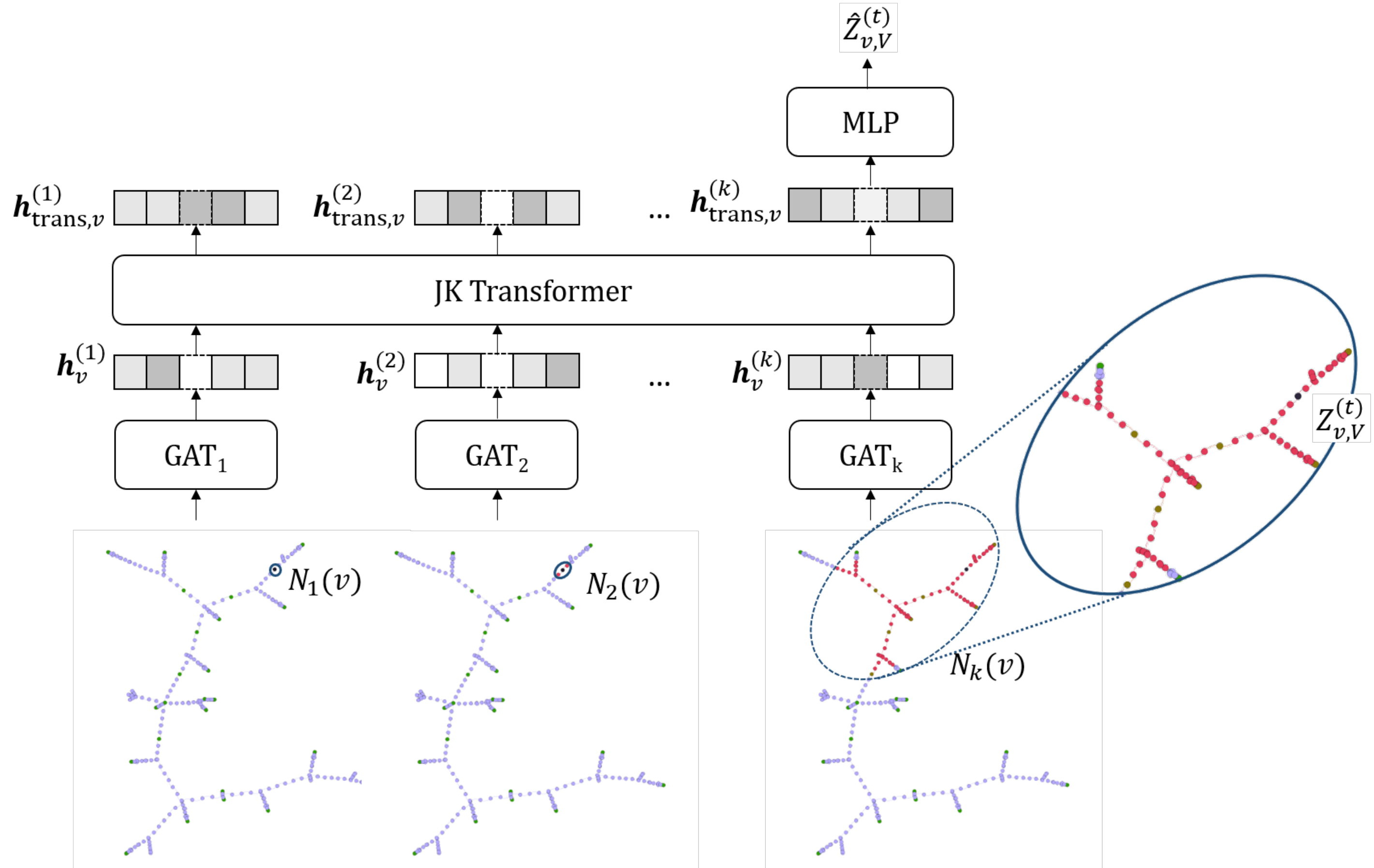}
 \caption{The Jumping Knowledge architecture with a Transformer. Red nodes are included in the receptive field $N_k(v)$ of the $k$-th GAT encoder for one example busbar $v$ here.}
 \label{fig:transformer_architecture}
\end{figure*}

\paragraph{Transformer encoder with Jumping Knowledge architecture}
To further investigate more efficient ways to deal with long-range dependence, we also propose another aggregation scheme for the JK architecture based on Transformer networks \cite{Vaswani2017}, which do not rely on backpropagation through time when encoding sequences as an alternative to LSTMs. We employ the Transformer's self-attention mechanism to the outputs of the stacked GNN layers (i.e. the intermediate node representations are treated like word embeddings in a machine translation model). After the self-attention unit, we also apply an add-and-normalize layer and a feed-forward network. Formally, in order to define the attention weights, we train three different linear weight matrices to calculate a query $\mathbf{Q}_{\text{Trans}}$ and a key $\mathbf{K}_{\text{Trans}}$ matrices of dimension $d_k$ and a value $\mathbf{V}_{\text{Trans}}$ matrix. The scores are calculated as described in \cite{Vaswani2017}: 
\begin{equation}
  \begin{split}
     \mathbf{Q}_{\text{Trans}} &= \mathbf{W}_{Q_{\text{Trans}}} \mathbf{h}_v^{(1)} \, \Vert \, \ldots \, \Vert \, \mathbf{h}_v^{(K)} \\
     \mathbf{K}_{\text{Trans}} &= \mathbf{W}_{K_{\text{Trans}}} \mathbf{h}_v^{(1)} \, \Vert \, \ldots \, \Vert \, \mathbf{h}_v^{(K)} \\
     \mathbf{V}_{\text{Trans}} &= \mathbf{W}_{V_{\text{Trans}}} \mathbf{h}_v^{(1)} \, \Vert \, \ldots \, \Vert \, \mathbf{h}_v^{(K)} \\
    \mathbf{H}_{\text{Trans}} &= softmax\left( \frac{\mathbf{Q}_{\text{Trans}}\mathbf{K}_{\text{Trans}}^T}{\sqrt{d_k}} \right)\mathbf{V}_{\text{Trans}}
  \end{split}
\end{equation}

In our particular semi-supervised node regression setting, we only need to feed the $K$ node representations of busbar nodes into the JK module, no matter if it is the LSTM-based or the Transformer encoding. We chose GAT to obtain the $K$ node embeddings $\mathbf{h}_k$ instead of the vanilla Graph Convolutional Network (GCN), since GAT uses the attention mechanism as a substitute for the statically normalized convolution operation of GCN. In our setting, attending over features of neighbors, instead of a simple mean aggregation is important since each node type has a different feature space and thus it may have a different relevance for the aggregation operation.

As shown in the overall architecture in Figure~\ref{fig:transformer_architecture}, the $K$ embeddings of a busbar node are encoded with the JK module and in case of the Transformer we pick the last output as a final representation for an MLP for the regression.


\section{Experiments}
\label{sec:experiments}
In this section, we introduce the details of our experiments to explore the effectiveness of GNNs for the task of state estimation.

\subsection{Experimental Setup}
\paragraph{Data Generation Process}
Our experiments are centered around a total of four different real-world power grid topologies that have been compiled in the context of the ENERA project\footnote{\url{https://projekt-enera.de}}, summarized in Table~\ref{tab:datasets}. All their equipment come with specified static attributes $\mathbf{C}$. Further, in each of the grids it is known where actual sensors are located. We denote the set of sensing equipment with at least one measurement by $\mathcal{V}_{\text{sensing}} = \left\{v_i \in \mathcal{V}| (i,j) \in \mathcal{I}_t \text{ for some } t,j \right\}$. Unfortunately, a large number of historical observations for dynamic states was missing. This is why we resort to a data generation procedure for two grids $\mathcal{G}^{(3)}$ and $\mathcal{G}^{(4)}$.
We obtain dynamic states at time $t_i$ by applying a scaling factor to the power level for each load in the grid based on an initial realistic load profile $t_0$. To get the (partially observed) dynamic states of equipment where sensors are located $Z_{\mathcal{I}_{t_i}}$ and ground truth labels for the busbar voltages $Z^{(t_i)}_{\mathcal{V}_\text{Busbar}, V}$ we execute a power flow solver on each of these new (scaled) load profiles ($t_i$). Since all the measurements and labels obtained this way are (unrealistically) perfectly aligned, we experiment with applying Gaussian noise $\mathcal{N}(0, p \hat{\sigma}^2)$ to the labels with a varying percentage $p$ of the estimated variance $\hat{\sigma} = Var(Z_{\mathcal{V}_{\text{Busbar}}, V})$. 

Throughout the evaluation, we use the mean-absolute error (MAE) as performance metric.

\begin{table}[t!]
\caption{Real-world power grid datasets used in experiments}
  \centering
  \begin{tabular}{c|c|c|c|c}
  \toprule
    Name & $|\mathcal{V}|$ & $|\mathcal{V}_\text{Busbar}|$ & $\frac{|\mathcal{V}_{\text{sensing}}|}{|\mathcal{V}|} $ & $|{t_i}|$ \\
    \midrule
    $\mathcal{G}^{(1)}$ & 298 & 32 & 8.7 \% & 8 \\
    $\mathcal{G}^{(2)}$ & 304 & 31 & 10.5 \% & 6 \\
    $\mathcal{G}^{(3)}$ & 282 & 28 & 8.5 \% & 42 \\
    $\mathcal{G}^{(4)}$ & 282 & 28 & 9.6 \% & 41 \\
    \bottomrule
  \end{tabular}
  \label{tab:datasets}
\end{table}

\begin{figure*}
 \centering
 \includegraphics[width=0.85\textwidth]{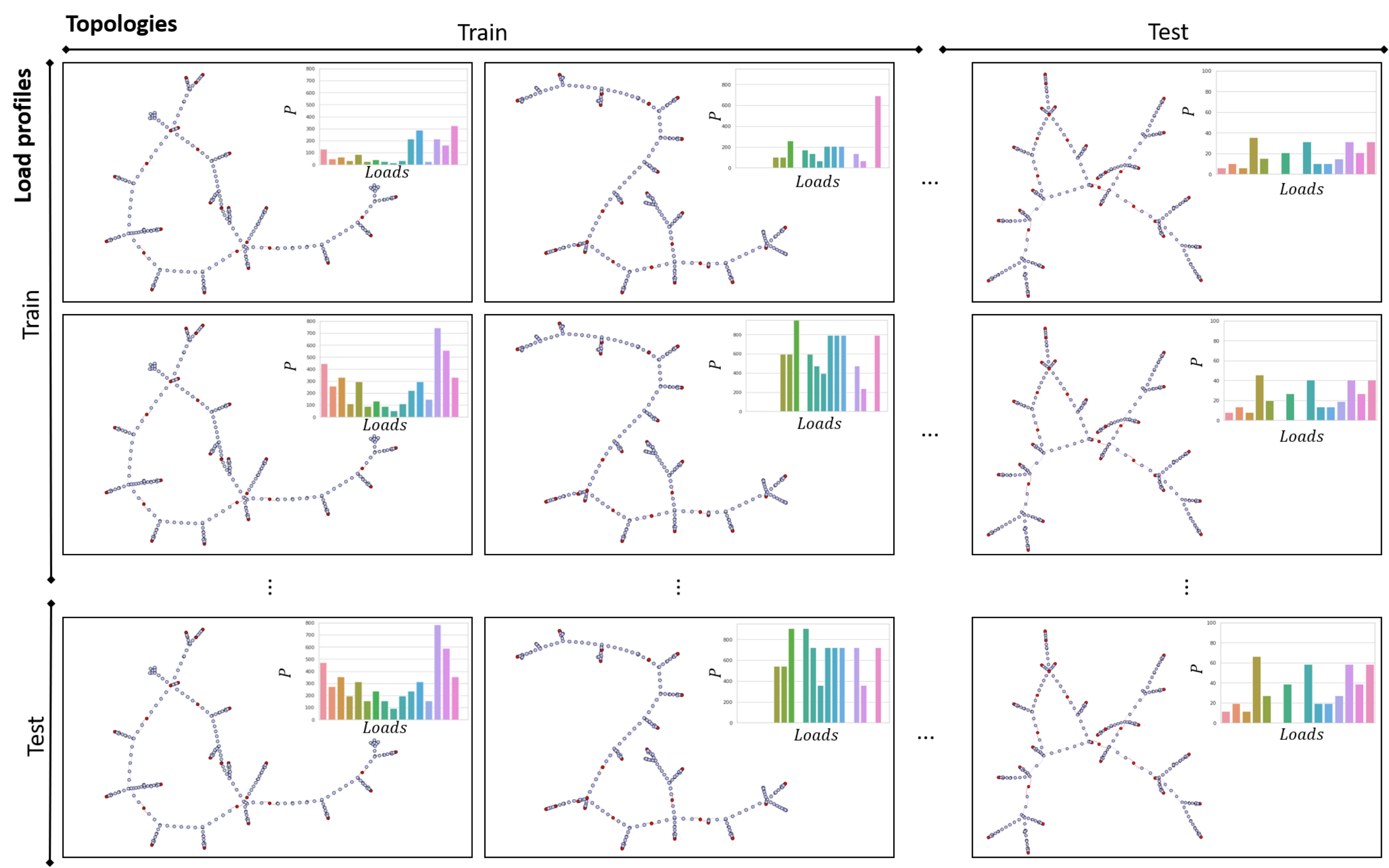}
 \caption{Two splitting strategies: vertical split: same topology different load profiles, horizontal split: different topology}
  \label{fig:splits}
\end{figure*}

\paragraph{Load Profile Split}
The first kind of generalization we want to investigate is concerning a single power grid $\mathcal{G}$ observed at different times $t$, i.e. operated at different load profiles. We define the dataset as a set of $m$ pairs of partially observed dynamic states (inputs) and busbar ground-truth labels (outputs):
$ \mathcal{D}_{\mathcal{G}} = \left\{ \left( Z_{\mathcal{I}_{t_1}}, Z^{(t_1)}_{\mathcal{V}_\text{Busbar}, V} \right), \left(Z_{\mathcal{I}_{t_2}}, Z^{(t_2)}_{\mathcal{V}_\text{Busbar}, V} \right), \allowbreak \dots, \left(Z_{\mathcal{I}_{t_m}}, Z^{(t_m)}_{\mathcal{V}_\text{Busbar}, V} \right) \right\}$.
To study generalization, we use a random (70/30) split of $\mathcal{D}$ into a training and a test set of load profiles. This split on load profiles is shown along the vertical axis in Figure~\ref{fig:splits}, where bar charts indicate the load profile distribution of active power for each load node respectively. As mentioned above, due to limited data, this setting only applies to $\mathcal{G}^{(3)}$ and $\mathcal{G}^{(4)}$.

\paragraph{Topology Split}
The second kind of generalization is concerning the inductive case across power grid topologies. We denote $\mathcal{G}^{(i)}$ as the $i$-th topology alongside its static features $\mathbf{C}^{(i)}$. We are interested to which degree a GNN that was trained on a set of topologies and load profiles can be transferred to a completely unseen topology (without any fine tuning).

Here we define a dataset of $l$ topologies observed with $m_l$ load profiles each, $\mathcal{D} = \left\{ \mathcal{D}_{\mathcal{G}^{(1)}}, \mathcal{D}_{\mathcal{G}^{(2)}}, \dots, \allowbreak \mathcal{D}_{\mathcal{G}^{(l)}} \right\}$.
We then perform a hold-one-out split with $l-1$ topologies in $\mathcal{D}$ as training and one hold out topology as test. 
This dataset split on load profiles is shown along the horizontal axis in Figure~\ref{fig:splits}.


\section{Results and Analysis}
\label{sec:results}

\begin{table}[]
\caption{Statistical Properties of datasets}
  \centering
  \begin{tabular}{l|l|c|c|c|c}
  \toprule
    Data & Domain & $\mu_{deg}$ & $dia$ & $\mu_{sp}$ & $\mu_{cc}$ \\
    \midrule
    MUTAG & Bio & 6.18 & 6 & 3.53 & 0.2 \\
    AIFB & KG & 5.59 & 8 & 3.88 & 0.015 \\
    Cora & Citation & 3.9 & 19 & 6.31 & 0.24 \\
    Citeseer & Citation & 2.74 & 28 & 9.32 & 0.14 \\
    Power grids & Electr. & 2.0 $(\pm 5e^{-3})$ & 67 ($\pm 15$) & 26.69 & 0 \\
    \bottomrule
  \end{tabular}
  \label{tab:statistical_analysis}
\end{table}

\begin{figure}[t]
 \centering
 \includegraphics[clip, trim=1cm 1cm 1cm 1cm,width=\linewidth]{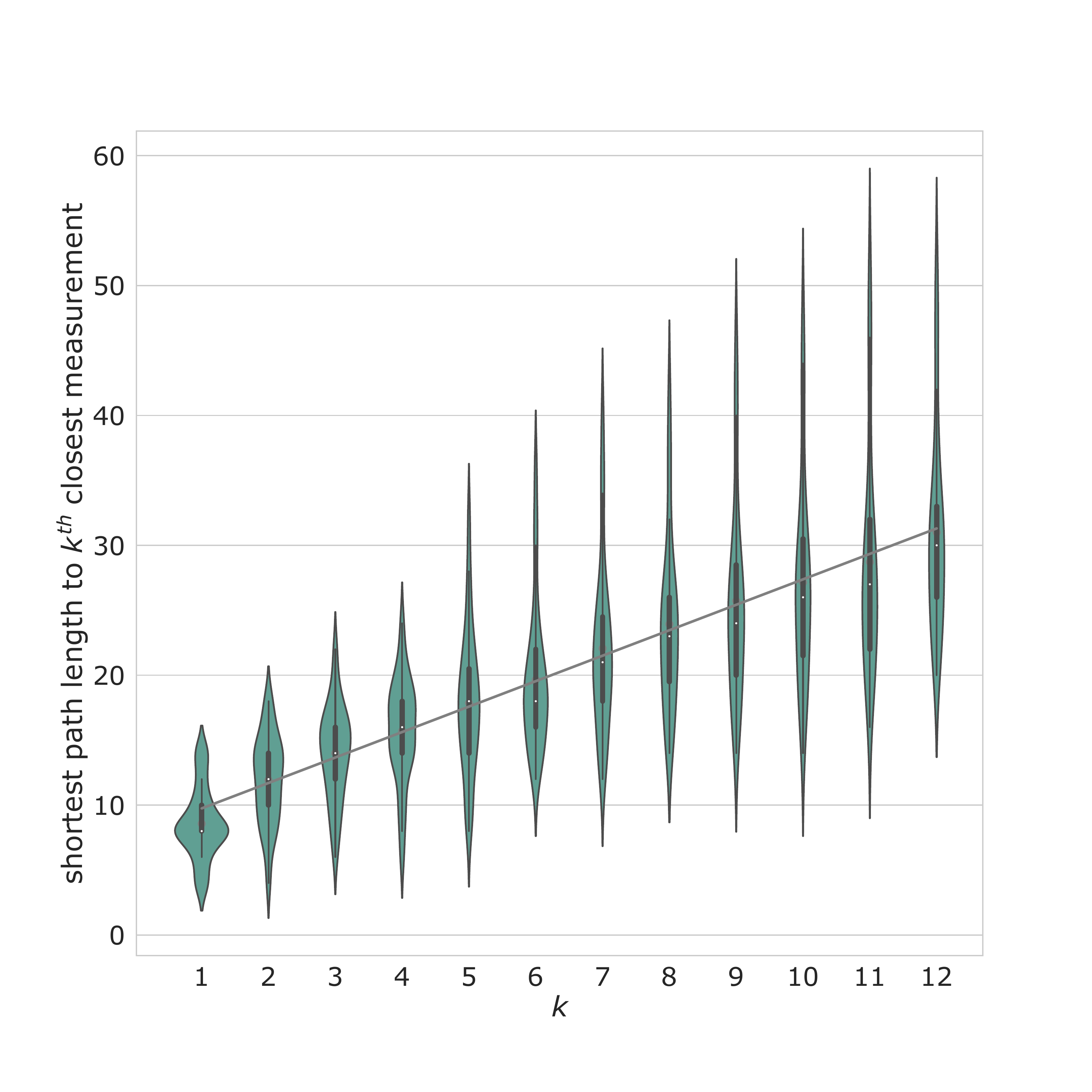}
 \caption{Distributions of shortest paths from any busbar to $k^{th}$ closest measurement node for all power grids}
  \label{fig:shortest_path_measurements}
\end{figure}

\subsection{Statistical Analysis of Data}
\paragraph{Power grids vs. traditional benchmarks}
Before going into the SE task results, we compare some key statistics of benchmark graph datasets to our real-world power grids. The statistics in Table~\ref{tab:statistical_analysis} summarize the comparison against the biological graph MUTAG, the university knowledge graph (KG) AIFB, and the citation graphs Cora and Citeseer. The rows are sorted w.r.t. the mean node degree $\mu_{deg}$. We further present the graph diameter ($dia$) (average for power grids), the average shortest-path length $\mu_{sp}$, and the average clustering coefficient $\mu_{cc}$. Values in parenthesis are the variances across all power grid graphs.

It can be seen that power grids are unique in all presented statistics. Especially the high mean shortest path length $\mu_{sp}$ between all pairs of nodes seems problematic. While benchmarks exhibit a small-world-like property - all nodes are maximally separated by ca. 9 hops, power flow may require the passing of very distant messages - up to 27 hops.
To emphasis this point for power grid data with sparse measurements, the violin plot in Figure~\ref{fig:shortest_path_measurements} shows the shortest path from busbar nodes to their $k$-th closest measurement node with increasing $k$. Even to the first closest measurement node, the shortest path has a mean length of $\approx8.5$ and seems to increase linearly instead of logarithmic as would be expected in a graph with a small-world property. Such long-range messages would not be passed in commonly used GNNs for benchmark datasets.

\paragraph{Load Profile Split}
As topology-agnostic baselines in the load profile split scenario, we consider a dummy regression model (Train Busbar Mean) which is simply predicting the mean of every busbar based on the training load profiles, a linear regression model (LR) and a basic two-layer NN (MLP). The LR model as well as the MLP get the flattened matrix $\mathbf{C}$ and zero-padded $\mathbf{Z}_t$ concatenated as a single vector input, while each busbar voltage in $Z^{(t_i)}_{\mathcal{V}_\text{Busbar}, V}$ corresponds to an output. Hence, the multi-output LR is actually $|\mathcal{V}_{\text{Busbar}}|$ independent models, each with global information, while the second layer of the MLP is not independent but shared for all busbar outputs.

\begin{figure}
 \centering
 \includegraphics[clip,trim=1.1cm 1.6cm 2.8cm 2.5cm,width=\linewidth]{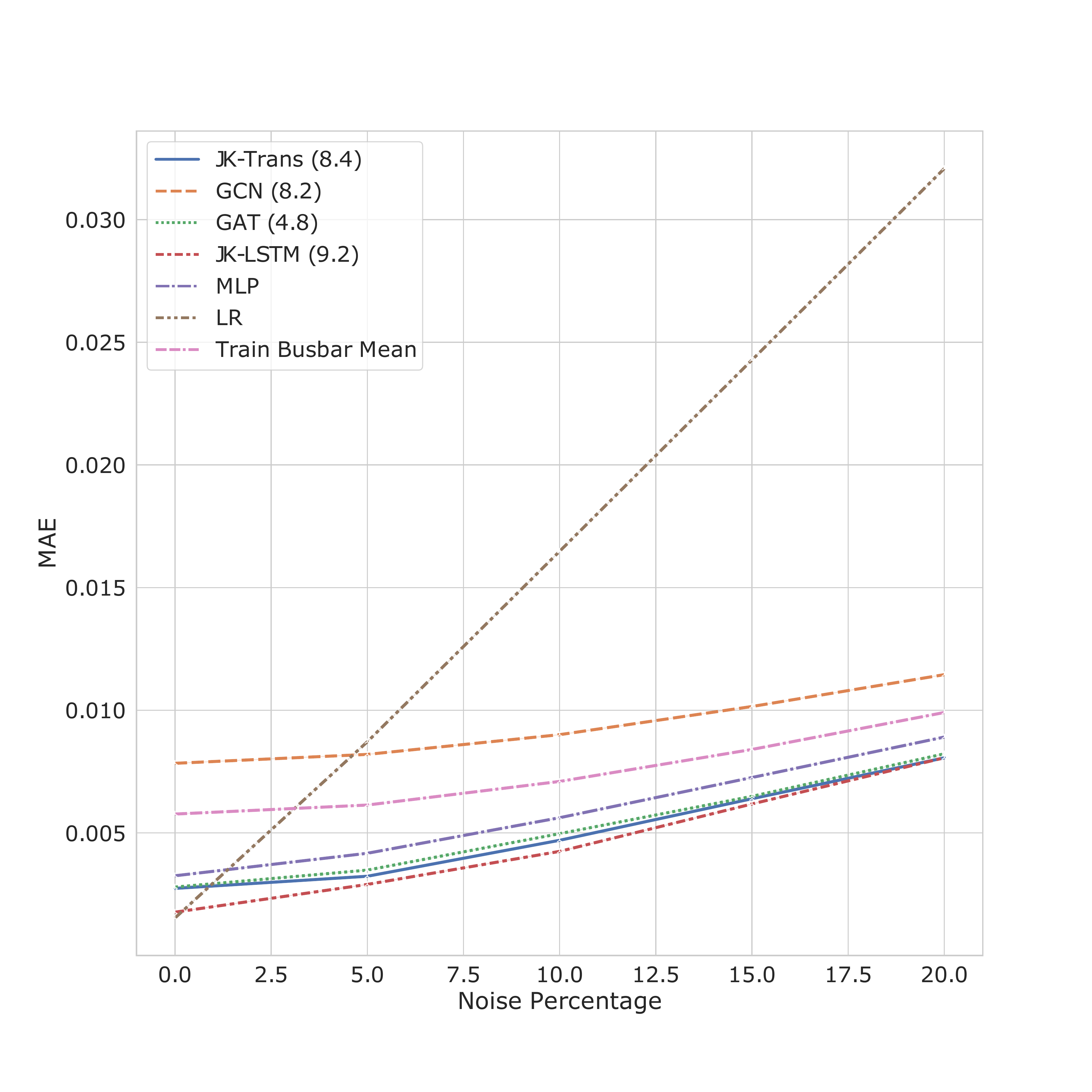}
 \caption{Performance on test load profiles for $\mathcal{G}^{(3)}$ w. increasing noise percentage. Mean number of layers for best models in brackets.}
  \label{fig:performance_load_profile_all}
\end{figure}
\begin{figure}
 \centering
 \includegraphics[clip,trim=1.1cm 1.6cm 2.8cm 2.5cm,width=\linewidth]{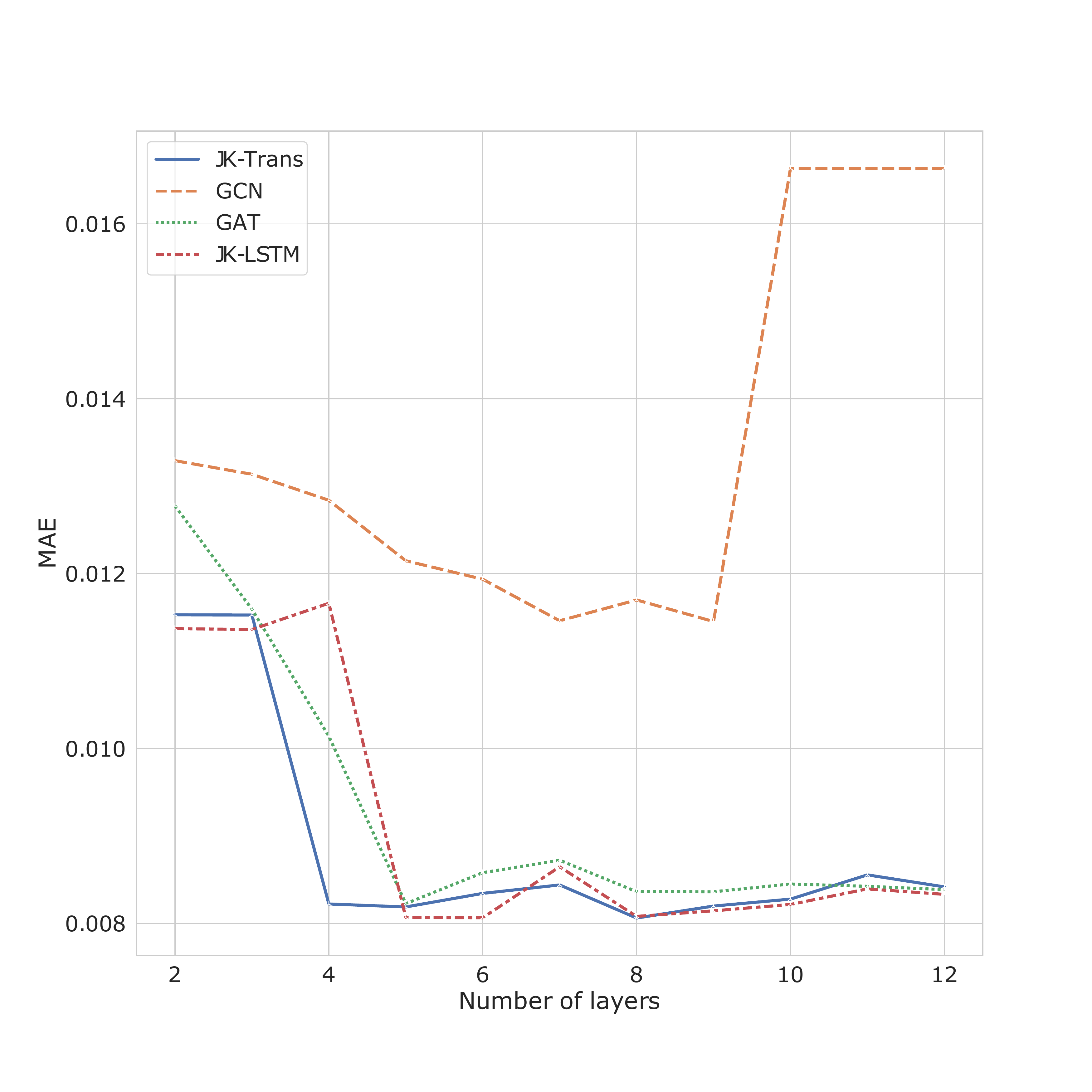}
 \caption{Load profile split performance for $\mathcal{G}^{(3)}$ w. 20\% label noise and increasing number of layers}
  \label{fig:performance_load_profile_GNN_only}
\end{figure}

For topology-aware models, we pick a vanilla GCN with mean aggregation, a standard stacking-based GAT, and the jumping knowledge architectures LSTM-based (JK-LSTM), and Transformer-based (JK-Trans). 
It can be seen that in the zero-noise case, the LR model outperforms all others. In this simple task, the relational inductive bias hinders the more complex models to fit as perfectly as the LR. However, with increasing noise the graph-based the LR's overfitting becomes evident and especially the JK models show less noise sensitivity. All graph-based models seem to follow the curve of the busbar mean baseline fitting a topology-aware mean. However, the GCN is the only model not able to beat this simple baseline.
The need for capturing long-range dependence is emphasized looking at Figure~\ref{fig:performance_load_profile_GNN_only}. There is a significant drop in error from layers 3 to 4 and 5. While the stacking GCN layers does suffer from over-smoothing, the GAT is almost on-par with the JK models. Table~\ref{tab:results_performance} summarizes the 0 and 20 \% noise scenarios and highlights the best models.

\begin{table*}[]
\caption{Summary of MAE performances for load profile splits. Relative error to best model as percentage in brackets.}
  \centering
  \begin{tabular}{c|c|c|c|c|c|c|c|c}
  \toprule
    Grid & Noise \% & JK-Trans & GCN & GAT & JK-LSTM & LR & MLP & Busbar Mean \\
    \midrule
    $\mathcal{G}^{(3)}$ & 0\% & 0.0027 (79) & 0.0078 (512) & 0.0028 (82) & 0.0018 (16) & \textbf{0.0015} (0) & 0.0033 (212) & 0.0058 (377) \\
    $\mathcal{G}^{(3)}$ & 20\% & \textbf{0.0081} (0) & 0.011 (42) & 0.0082 (2) & 0.0081 (0.04) & 0.0321 (398) & 0.0089 (11) & 0.0099 (23) \\
    $\mathcal{G}^{(4)}$ & 0\% & 0.0071 ($\approx118k$) & 0.0271 ($\approx451k$) & 0.0059 ($\approx98k$) & 0.0028 ($\approx46k$) & $\mathbf{6e^{-6}}$ (0) & 0.0152 ($\approx253k$) & 0.0228 ($\approx379k$) \\
    $\mathcal{G}^{(4)}$ & 20\% & 0.0114 (44) & 0.0263 (333) & 0.0107 (36) & \textbf{0.0080} (0) & 0.1007 (1272) & 0.0189 (238) & 0.0239 (302) \\

    \bottomrule
  \end{tabular}
  \label{tab:results_performance}
\end{table*}

\paragraph{Topology Split}
In the inductive topology split scenario, the results are all based on a hold-one-out test topology strategy. None of the topology-agnostic baselines would be applicable in this setting. Hence, we focus on the GNN based architectures: GCN, GAT, and JK models. As a very simple baseline, however, we do obtain the global mean over all busbar labels in the training topologies and apply it to the hold-out one. To separate the de-noising behavior from the transfer, we are not adding noise to the labels for this inductive setting.

\begin{figure}[t]
 \centering
 \includegraphics[clip,trim=0.85cm 1.6cm 2.8cm 2.5cm,width=\linewidth]{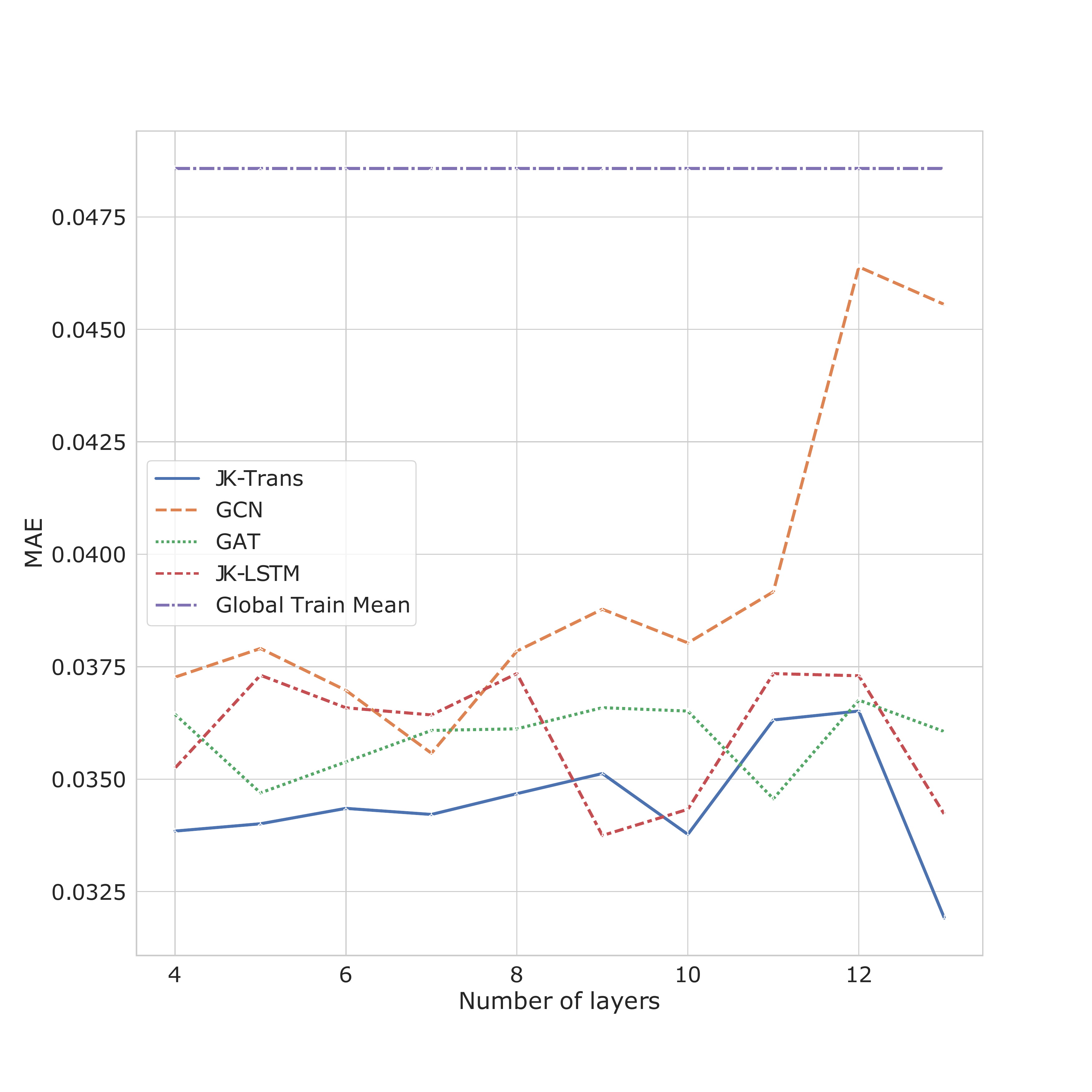}
 \caption{Inductive setting performance on hold-out topology $\mathcal{G}^{(4)}$ with increasing number of layers}
  \label{fig:performance_topologies}
\end{figure}
In Figure~\ref{fig:performance_topologies}, we report the results when training on $\mathcal{G}^{(1)},\mathcal{G}^{(2)},\mathcal{G}^{(3)}$ holding out topology $\mathcal{G}^{(4)}$ and its test load profiles. This is a difficult transfer scenario that has not been studied so far. It can be seen that the there is a transfer effect for GNNs to this hold-out topology as they all outperform the simple baseline. Interestingly, the JK-Trans outperforms all other GNN models. There is a consistent drop happening from layer 12 to 13 and lowest error for JK-Trans is achieved using 13 layers with a significant drop. We did not include higher number of layers in our hyperparameters, but there might be even room for improvement beyond 13. 


\begin{table}[]
\caption{Performances for hold-out topology $\mathcal{G}^{(4)}$}
  \centering
  \begin{tabular}{c|c|c}
  \toprule
    Model & Performance & Relative error \\
    \midrule
    JK-Trans & 0.0319 & 0 \% \\
    JK-LSTM & 0.0337 & 5.6 \% \\
    GAT & 0.0346 & 8.5 \% \\
    GCN & 0.0355 & 11.3 \% \\
    Global Mean & 0.0486 & 52.4 \% \\
    \bottomrule
  \end{tabular}
  \label{tab:results_performance}
\end{table}



\section{Discussion}
\label{sec:discussion}
Our results show that making ML models for SE topology-aware has benefits when it comes to handling noisy labels. We attribute this to the GNNs behavior of acting like a low-pass filter on high-frequency graph signals. In the unrealistic scenario with perfect measurements and ground truth, the SE task is too simple and more complex ML models with shared layers are not suitable. 

In the inductive setting, our results show that the JK-Trans model is a first step towards satisfying performance. However, in order to make GNNs work as surrogates for power flow solvers, still more efficient architectures are needed. 

With these encouraging first results, we want call the ML and data science community's attention to the issue that progress in GNN research in this domain is still lacking standardized and well-understood datasets. Together with power system domain experts we will actively engage in the contribution of our ENERA power grids in a graph data model suitable for inductive learning to the OGB project \cite{hu2020open}. 
A good role-model is the current flourishing of GNN research on biomedical KGs due to standardized benchmarks and well-defined tasks such as drug discovery \cite{himmelstein2017,liu2020}.

\section{Related Work}
\paragraph{Power Grid State Estimation}
For SE in practice, commonly used approaches are weighted least squares models, which suffer from their iterative global fitting procedure and become computationally very expensive for larger systems\cite{Zhang2019}. To model non-linear dependencies and improve the scalability, several neural network models have been proposed, from simple feed-forward networks to deep recurrent networks \cite{Cao2020}. These models predict surrogate states given $k$ previous ground-truth states in a timeseries using smoothness of signals assumption across time instead of spatial smoothness exploited by GNNs. None of these approaches are capable of inductive learning across different of topologies.

\paragraph{Graph Representation Learning and Power Grids}
Some early work using graphical models for power grids \cite{Weng2013}, where the inference procedure is based on message-passing is conceptually similar to GNNs. The major difference is that the graphical model explicitly models the non-linear power equations in a probabilistic way (depending on the topology), while GNNs learn to encode non-linear messages using a set of trainable weights.

Recently, the authors in \cite{ChenHZYH20} presented a GNN model for fault localization in a simple distribution system which consists of 128 busbar nodes with simulated loads and faults. Even in this simplified setting, they recognize the need to deal with unconventional long-range dependence. They perform a data augmentation on the topology, adding shortcut edges from the $k$-nearest neighbors graph ($k = 20$). Instead of a vanilla GCN, they also use the $K$-th order ($K$-hop) Chebyshev polynomials of the Laplacian for each graph convolutional layer. The best results are obtained with a GNN with 3, 4, and 5-th order Laplacian for the 3 GCN layers, respectively.
Essentially, their proposed GNN's supportive field is the full graph (every node receives messages from every other node) and the model loses the inductive learning capability, since the learned filters only apply to the spectral decomposition of one topology.


The work of Donon et al. in \cite{Donon2019} is the only GNN model to our knowledge that studies transferability across power grid topologies. It was shown that GNNs can effectively approximate power flow solvers and generalize to novel topologies in transmission systems. Although the authors operate on simplified simulated topologies, they show the need for a dedicated GNN for this task which is unlike commonly known ones. Our empirical study shares similarities in the ML task definition, but the setup for our real-world distribution systems is much more complex, as we discussed in the graph representation section.



\section{Conclusion}
The increasing uncertainties and dynamic operational conditions of today's smart power grids require more robust and fast monitoring tools. Traditional power flow solvers are slow and cannot handle noisy measurements. On the other hand, topology-bound ML approaches are not flexible enough and need to be re-trained even in case of small topology changes. Motivated by these limitations, we studied the application of topology-aware GNNs for state estimation in real-world power grids.
While initiatives to collect diverse graph benchmark datasets for GNN research exist, they lack power grids entirely. Due to the practical relevance of power grids and the fact that they differ significantly from other well-known graphs (e.g., in terms of structural properties and sparse features and labels), we believe that more attention is needed to explore the opportunities for GNNs in this domain.

We propose a more expressive graph representation of power grids suitable for inductive learning across different topologies.
Interestingly, we find that GNNs are significantly more robust to noisy measurements with up to 400\% lower error compared to simple baselines.
Our results further show that the typical over-smoothing effect of GNNs is not as immediate and find the best performing models to be exceptionally deep with up to 13 layers. Hence, GNNs in this field need rather special, carefully designed architectures to effectively handle long-range dependence.

\begin{appendix}

\section{Appendix}
\begin{figure}[t!]
 \centering
 \includegraphics[clip,trim=0.7cm 1.5cm 2cm 2cm, width=0.8\linewidth]{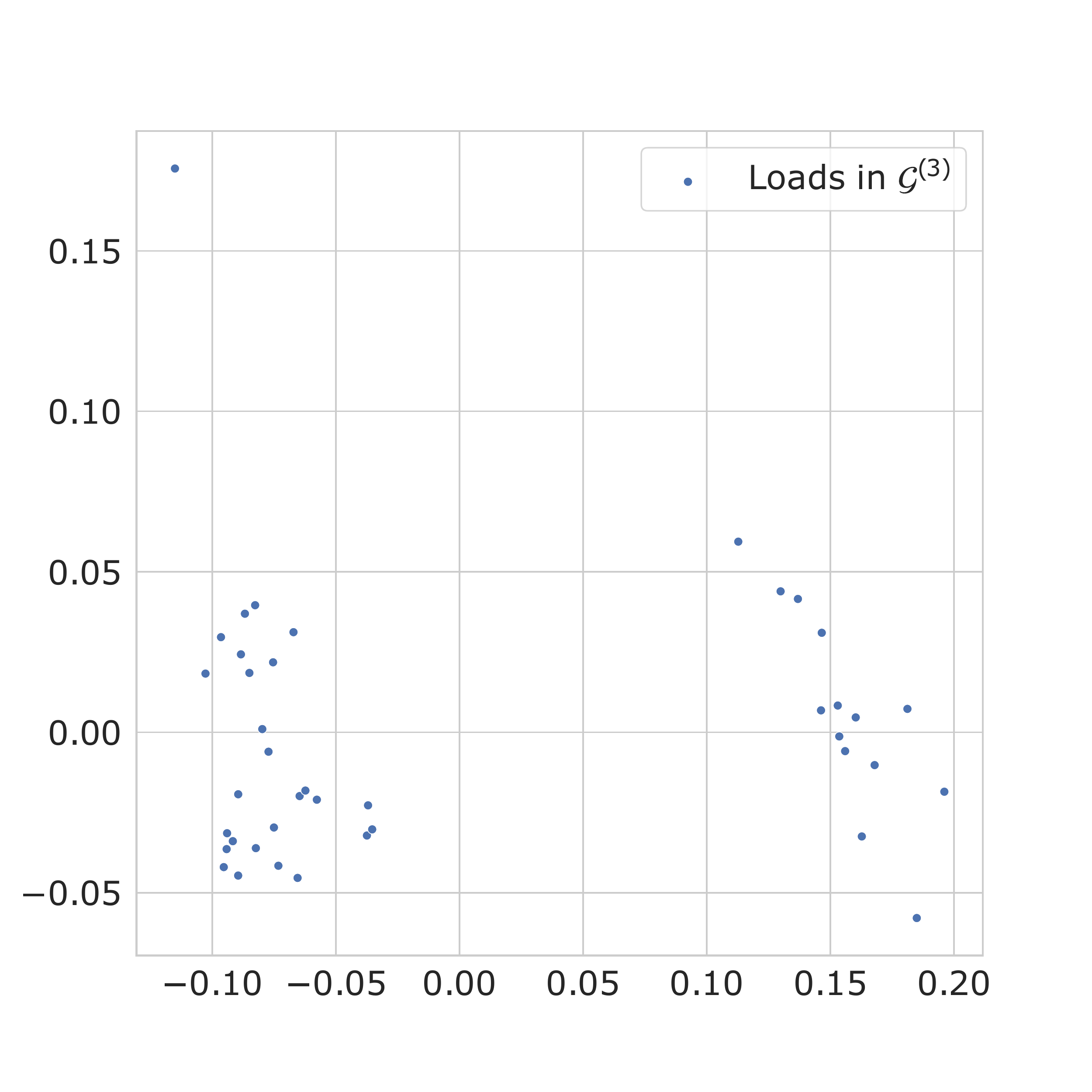}
 \caption{Load profile distributions across time projected to 2-dimensions using PCA}
 \label{fig:loads}
\end{figure}

This appendix gives more details on the used node features of our power grids and the load profile data generation process. 
We investigate the distribution of load profiles obtained with our scaling procedure by projecting all loads across time to two dimensions using principal component analysis. In Figure~\ref{fig:loads} each point corresponds to a load node in $\mathcal{G}^{(3)}$ with PCA applied across time.
There are two high-level load profile clusters and one outlier profile. Random splits on these profiles should be acceptable as it will most likely yield profiles from each of the clusters in train and in test.
We also observed that feature standardization to zero-mean and unit variance is important, since the features lie on different scales. Gradient-based optimization is susceptible to being dominated by features on the larger scales. Further, we apply min-max scaling to the busbar voltage labels based on the respective training dataset.


We used following ranges of hyperparameters for the GNN models:
Number of layers $\in \left \{2,3,4,5,6,7,8,9,10,11,12,13\right \}$, attention heads $\in \left\{1,2 \right\}$, learning rate $\in \left\{1e^{-3}, 1e^{-4}, 1e^{-5} \right\}$, $d_k \in \left\{32, 64\right\}$, dropout for all layers $p_{do} \in \left\{0.0, 0.1\right\}$.
For aggregation the vanilla GCN uses a mean, the GAT a summation.
\end{appendix}

\bibliographystyle{ACM-Reference-Format}
\balance
\bibliography{library}

\newpage

\end{document}